\documentclass[runningheads]{llncs}

\usepackage{eccv}

\usepackage{eccvabbrv}

\usepackage{amsmath}
\usepackage{amssymb}
\usepackage{graphicx}
\usepackage{subcaption}
\usepackage{tabularx} \newcolumntype{C}{>{\centering\arraybackslash}X} \newcolumntype{L}{>{\raggedright\arraybackslash}X} \newcolumntype{R}{>{\raggedleft\arraybackslash}X}
\usepackage{booktabs}
\usepackage{multirow}
\usepackage{multicol}
\usepackage{dsfont}
\usepackage{pifont} \newcommand{\xmark}{\ding{55}}

\usepackage[accsupp]{axessibility}  

\usepackage[pagebackref,breaklinks,colorlinks,citecolor=eccvblue]{hyperref}

\usepackage{orcidlink}

\begin{document}

\title{PuzLM: Solving Jigsaw Puzzles with Sequence-to-Sequence Language Models}

\titlerunning{PuzLM: Sequence-to-Sequence  Jigsaw Puzzle Soslving}

\author{Gur Elkin \and
Ofir Itzhak Shahar \and
Ohad Ben-Shahar}

\authorrunning{G. Elkin et al.}

\institute{Ben-Gurion University of the Negev}

\maketitle

\begin{abstract}
    Square jigsaw puzzles are typically solved by visually matching piece images to recover the original layout. This work introduces \textbf{PuzLM}, an alternative perspective that recasts jigsaw reassembly as a discrete sequence-to-sequence (Seq2Seq) problem, inspired by natural language representations. We design an efficient puzzle quantization procedure that transforms each piece into a short sequence of discrete tokens, enabling the direct application of standard Seq2Seq language models as powerful jigsaw solvers. Our approach demonstrates that accurate puzzle reconstruction can be achieved through purely symbolic reasoning over discrete representations, improving state-of-the-art performance even on puzzles with eroded boundaries or missing pieces.
    \keywords{Jigsaw Puzzles \and Seq2Seq \and Image Tokenization}
\end{abstract}

\begin{figure}[b]
    \centering
    \includegraphics[width=\linewidth]{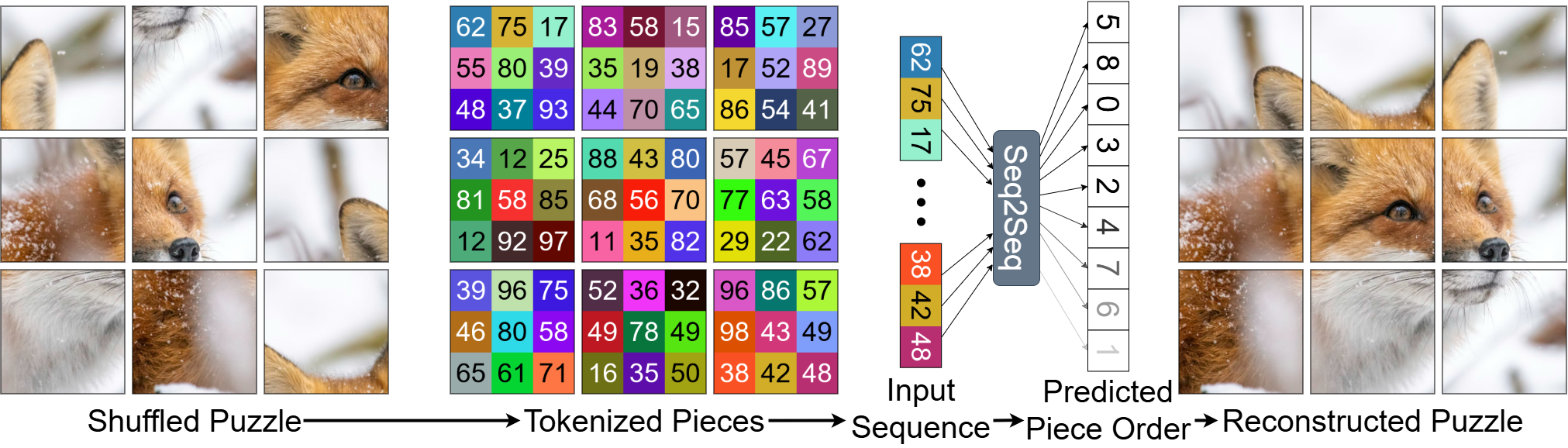}
    \caption{Overview. Each puzzle piece is converted into a short sequence of discrete tokens using a lightweight quantization process. Jigsaw reassembly is then formulated as a sequence-to-sequence prediction problem, where a standard Seq2Seq model autoregressively predicts the global ordering of pieces \textit{using only symbolic inputs}.}
    \label{fig:overview}
\end{figure}

\section{Introduction}
\label{sec:intro}
Although jigsaw puzzles are a popular pastime for humans, their algorithmic reassembly remains an enduring computational challenge, with important applications such as reconstructing shredded documents, reassembling broken artifacts and mosaicing satellite imagery~\cite{rika2025generic,tsesmelis2024re,soille2006morphological}. Square jigsaws, created by partitioning an image into equal-sized tiles, are regarded as one of the most fundamental yet challenging instances of the problem. Unlike commercial toy puzzles with uniquely shaped pieces, square puzzles lack geometric cues that could guide reconstruction, forcing solvers to rely primarily on their visual content.

Consequently, most existing approaches reason about jigsaw reassembly by directly accessing raw pixels or learned visual embeddings of individual pieces~\cite{pomeranz2011fully,sholomon2013genetic,paikin2015solving,elkin2025recognizing}. Within this paradigm, visual appearance and structural patterns (such as relative ordering, global layout consistency, or long-range dependencies between distant pieces) are jointly represented in a single, continuous feature space. While effective, this coupling makes it difficult to distinguish whether successful reconstruction arises from local visual cues or from higher-level structural patterns that emerge only at the level of the assembled puzzle. As a result, the role of non-explicit, relational information remains largely unexplored.

In this work, we seek to explicitly disentangle structural reasoning from direct visual access in jigsaw reassembly. Our objective is not to replace visual methods, but to investigate whether accurate reconstruction can be achieved by leveraging structural patterns alone, which are independent of raw pixel information. To this end, we introduce a symbolic representation of puzzle pieces and reformulate square jigsaw reassembly as a Seq2Seq prediction problem, operating exclusively on discrete tokens. This design enables a principled study of purely symbolic puzzle solving while still capturing global structural regularities. Our research offers both conceptual and functional contributions:
\begin{enumerate}
    \item \textbf{Symbolic reformulation of jigsaw reassembly.} We recast square jigsaw puzzle reconstruction as a Seq2Seq prediction problem by introducing a discrete, symbolic representation of puzzle pieces, offering an alternative to pixel- and feature-based approaches.
    \item \textbf{Efficient piece quantization scheme.} We propose a lightweight quantization procedure that converts each puzzle piece image into a short sequence of discrete tokens, abstracting visual variability while preserving the structural cues required for reconstruction.
    \item \textbf{Leveraging Seq2Seq language models as jigsaw solvers.} We demonstrate how standard encoder–decoder Seq2Seq architectures can be directly applied to puzzle solving, without requiring vision-specific adaptations.
    \item \textbf{Reassembly without continuous access to visual input.} Once the raw visual input is tokenized, the solver no longer accesses the former and operates exclusively on the tokenized representations. We show that symbolic patterns alone, independent of pixel-level cues, are sufficient to guide accurate puzzle reconstruction.
    \item \textbf{Robust performance.} Across multiple benchmarks, our approach achieves competitive or superior accuracy to recent state-of-the-art (SOTA) vision-based solvers, including challenging settings such as eroded or missing pieces.
\end{enumerate}

\section{Related Work}
\label{sec:rel_work}

Jigsaw puzzle reassembly lies at the intersection of computer vision, combinatorial optimization, and spatial reasoning. Since its inception as a computational problem~\cite{freeman2006apictorial}, researchers have studied puzzles of various shapes and scales~\cite{harel2024pictorial,derech2021solving,shahar2025pairwise,gur2017square}, with square jigsaws occupying a particularly significant role. Because all pieces share identical geometry, reconstruction must rely almost entirely on visual cues, leading prior work to focus predominantly on extracting and comparing pictorial features. The lines of research surveyed below illustrate the depth of this paradigm and provide context for our alternative perspective.

\textbf{Square Jigsaw Puzzle Solving.}
Computational approaches to jigsaw puzzle reassembly have a long history, with early methods focusing on defining pairwise compatibility metrics between pieces and incorporating them into an auxiliary optimization framework~\cite{pomeranz2011fully,paikin2015solving,gallagher2012jigsaw,sholomon2013genetic,vardi2023multi}. With the advent of deep learning, subsequent works naturally replaced hand-crafted features with learned compatibility functions based on convolutional neural networks~\cite{sholomon2016dnn,paumard2020deepzzle,li2021jigsawgan}. More recently, the rise of generative modeling in computer vision motivated the adoption of this paradigm for jigsaw puzzle solving. Generative adversarial networks and diffusion models were used to inpaint missing and eroded pieces~\cite{talon2022ganzzle,talon2025ganzzle++,liu2024solving}, while spatial diffusion approaches progressively ``denoise'' scrambled pieces into coherent solutions~\cite{scarpellini2024diffassemble}. A complementary line of research introduced compatibility-based reward functions within reinforcement learning frameworks to guide reconstruction~\cite{song2023siamese,song2023solving,song2025erlmpp}. Other studies leveraged pretrained Vision Transformers (ViTs), relying on their strong visual feature extraction capabilities~\cite{heck2025solving,kim2025solving}. Finally, recent works have begun to incorporate multimodal signals, such as textual descriptions~\cite{xu2025vlhsa}, and to explore the use of large vision-language models (LVLMs) for jigsaw solving, though under constrained settings~\cite{lyu2025jigsaw,wang2025jigsaw}.

\textbf{Jigsaw Understanding as a Pretext Task.}
Jigsaw puzzles have also been widely used as a self-supervised pretext task for representation learning, where models predict the correct permutation of shuffled image patches~\cite{noroozi2016unsupervised,kolesnikov2019revisiting}.In this setting, reconstruction serves as a means for learning transferable visual features rather than as the primary objective. In contrast, and like the methods above, our work focuses directly on accurate puzzle reassembly.

\textbf{Vector Quantization and Tokenized Image Representations.}
Discrete and tokenized image representations have emerged as a powerful abstraction for modeling visual data. VQ-VAE~\cite{van2017neural} demonstrated that images can be represented as discrete codes drawn from a learned codebook, enabling the application of sequence models to image synthesis and layout generation~\cite{esser2021taming}. More broadly, tokenization is used to reduce visual variability, improve robustness, increase efficiency and bridge vision with models originally developed for discrete domains such as natural language processing (NLP)~\cite{chang2022maskgit,yu2024image}. Our puzzle-piece tokenizer is inspired by this paradigm, but deliberately designed to preserve relational structure relevant for reassembly rather than, \eg, promote visual fidelity.

\textbf{Sequence Models for Spatial Reasoning.}
The success of transformer-based sequence models~\cite{vaswani2017attention} in NLP motivated their adoption to other domains, such as vision. Encoder-only ViT~\cite{dosovitskiy2020image} initially demonstrated the potential of these architectures for image classification, and is now widely used for representation learning~\cite{simeoni2025dinov3}. Encoder-decoder transformers have been successfully applied to numerous problems such as object detection~\cite{carion2020end} and image generation~\cite{chang2022maskgit}. Our work contributes to this trend by showing that jigsaw puzzle reassembly, a combinatorially hard spatial reasoning problem, can be effectively formulated as a Seq2Seq task for standard encoder-decoder transformers.

Taken together, these threads reveal a gap in the existing literature. While jigsaw solvers have become increasingly sophisticated, they remain grounded in direct visual input. At the same time, sequence models and tokenized representations have proven effective for structured prediction across diverse domains. Our work bridges these directions by investigating whether square puzzles can be reconstructed using a standard Seq2Seq model operating solely on symbolic abstractions, thereby decoupling structural reasoning from direct visual access.

\section{PuzLM}
\label{sec:method}

An essential observation for our research is the similarity between square jigsaw puzzle solving and various Seq2Seq NLP tasks such as machine translation: given a sequence from the source distribution (\eg, text in French or shuffled puzzle pieces), we want to predict its corresponding sequence in another distribution (\eg, text in English or reconstructed puzzle positions). The following sections detail the computational process that enables us to effectively map puzzle reassembly to the discrete sequence modeling domain.

\subsection{Problem Formulation}

Square jigsaw puzzles over an $n\times n$ grid (for $1<n\in\mathbb{N}$) are typically given as an unordered set of piece images $P=\{p_1,\dots,p_{n^2}\}\subset\mathbb{R}^{H\times W\times C}$, where each piece $p_i$ has height $H$, width $W$, and $C$ color channels. To simplify the notation, let $N=n^2$, and define zero-based raster order over the grid such that position $[i,j]$ corresponds to a single index $i\cdot n+j$. A valid solution is a bijective mapping $Y:P\to\{0,\dots,N-1\}$ that assigns each piece to a unique grid position.

Since most puzzle solvers process inputs in some order (even if arbitrary), it is widely accepted to implement $Y$ as a permutation over the indices $\{0,\dots,N-1\}$. In this formulation, $Y[i]=j$ indicates that the $i$\textsuperscript{th} input piece should be placed at grid position $j$. A learning-based jigsaw puzzle solver is therefore a parametric function $f_\theta:\mathbb{R}^{N\times H\times W\times C}\to\{0,....,N-1\}^N$. During training, the solver is optimized over a dataset of puzzle-solution pairs $\{(P^{(i)},Y^{(i)})\}_{i=1}^M$ to maximize agreement between the predicted and ground-truth assignments.

In our formulation, the puzzle pieces are first tokenized using a quantization function $q_\phi:\mathbb{R}^{N\times H\times W\times C}\to V^{T(N)}$, where $V=\{0,\dots,|V|-1\}$ denotes a finite vocabulary of discrete tokens. The term $T(N):\mathbb{N}\to\mathbb{N}$ specifies the total number of tokens produced for a puzzle of size $N$. In practice, $T$ scales linearly with $N$ and depends on the number of tokens extracted per piece, which we refer to as the \emph{granularity} of $q_\phi$. The quantizer may be either fixed or parameterized by $\phi$, as is common in modern image tokenizers.
This representation allows our solver to operate end-to-end in the discrete domain. Thus, we define $s_\theta:V^{T(N)}\to\{0,\dots,N-1\}^N$, which predicts a permutation over grid positions from the token sequence. The overall puzzle-solving pipeline is thus a composition $f_{\phi,\theta}=s_\theta(q_\phi(P))$, i.e., tokenization followed by discrete permutation prediction.

\subsection{Puzzle Tokenization}

\begin{figure*}[t]
    \centering
    \includegraphics[width=\linewidth]{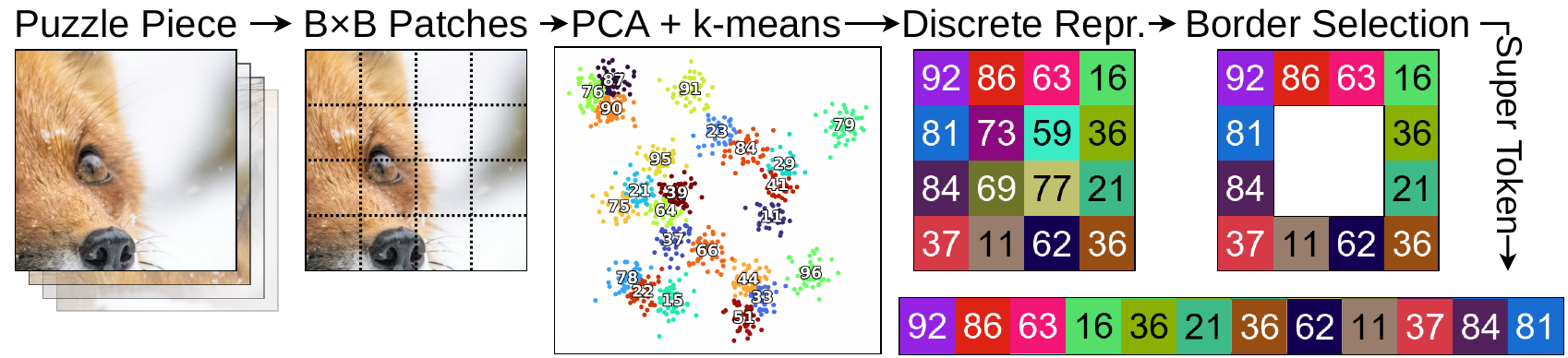}
    \caption{Tokenization Process. Each of the $N$ shuffled pieces is divided into $B \times B$ \textit{patches} (above $B=4$). Next, all patches are projected to a lower-dimensional space via a PCA matrix. We then associate each patch with the index of its nearest centroid (through $k$-means clustering). Lastly, for each piece we retain only the $b = \max(1, 4B-4)$ patches that lie on its border, chaining them clockwise into a \textit{super-token}. The full puzzle is then represented as the concatenation of all $N$ super-tokens.}
    \label{fig:tokenization}
\end{figure*}

The goals of our proposed piece quantization procedure are twofold:
\begin{enumerate}
    \item \textbf{Structural Coherence.} Since puzzle solving fundamentally relies on non-trivial structural pattern matching between pieces, as well as enforcing global coherence, each token should correspond to a well-defined pattern. In particular, token identities should be invariant to their position within the sequence and should not depend on surrounding tokens. This ensures that matching decisions are driven by intrinsic structure rather than encoding artifacts.
    \item \textbf{Computational Efficiency.} Because puzzle tokenization is an integral first step before running our solver, it must introduce minimal computational overhead. This would support fast, on-the-fly processing of previously unseen puzzles, as well as scalable preprocessing of large datasets for training.
\end{enumerate}
Our proposed jigsaw-oriented tokenizer thus performs unsupervised quantization over the training pieces  in the following way (visualized in \cref{fig:tokenization}):
\begin{enumerate}
    \item \textbf{Patch Extraction.} Each puzzle piece is divided into a $B \times B$ grid of \textit{patches}, where $1\le B\in\mathbb{N}$ controls the tokenizer's granularity.
    
    \item \textbf{Dimensionality Reduction.} The extracted patches are flattened and projected to a lower-dimensional space (denoted $\mathbb{R}^d$) via a Principal Component Analysis (PCA) matrix. This was shown to improve clustering efficiency by reducing noise from irrelevant visual variance~\cite{mukherjee2024capturing}. 
    
    \item \textbf{Clustering.} We apply $k$-means clustering to the projected patches. Then, each patch is represented by the \textit{index} of its nearest centroid. The choice of $k$ defines the tokenizer's vocabulary size $|V|$, with higher values offering a finer distinction between similar vectors.
    
    \item \textbf{Border Selection.} We retain the $b = \max(1, 4B-4)$ tokens located on the borders of each piece. These are concatenated in clockwise order to form a \textit{super-token}: a short discrete sequence that represents one piece. This step reduces sequence length (from quadratic to linear in $B$), while preserving the most relevant information for puzzle solving.
\end{enumerate}

This process yields a compact, discrete, and spatially-aware representation of puzzle pieces. Most importantly, the meaning of each token is independent of its relative position within the super-token, while fixed ordering ensures that the spatial configuration of border tokens remains consistent across pieces. Unlike deep image quantizers~\cite{van2017neural,yu2024image}, which often entangle patch representations and require complex decoding schemes, our method offers a transparent, efficient, and easily scalable solution (see comparison in \cref{tab:abl-tokenizer}).

Once all pieces are converted into super-tokens, we concatenate them into a single input sequence to represent the entire puzzle. Since the original set is unordered, we impose a lexicographic order over the super-tokens to give the input a consistent structure. Additionally, we insert a dedicated separator token \text{\texttt{[SEP]}}, between every pair of super-tokens. This makes their bounds explicit (analogous to spaces between words), encouraging the model to treat each piece's super-token as a coherent semantic unit. Overall, we have found our design choices promote accurate reassembly (see ablations in \cref{tab:abl-tokenizer}).

With the steps outlined above, given a puzzle $P\in\mathbb{R}^{N\times H\times W\times C}$, our tokenizer $q_\phi$ thus transforms it into a discrete sequence:
\begin{equation}
X=\left(
x_1^1,\dots,x_1^b,
\text{\texttt{[SEP]}},
x_2^1,\dots,x_2^b,
\text{\texttt{[SEP]}},
\dots,
\text{\texttt{[SEP]}},
x_N^1 \dots,x_N^b
\right)
\end{equation}
where $x_i^1,\dots,x_i^b$
represents the $i^{\mathrm{th}}$ piece's super-token. Hence, the overall sequence length is $T(N)=bN+(N-1)$.

\subsection{Discrete Jigsaw Representations}

We aim to solve jigsaw puzzles by reasoning over symbolic, discrete representations rather than continuous visual features. This approach is orthogonal to most contemporary solvers, which extract deep features directly from pixel intensities. Our goal is not to argue that discrete representations are universally superior, but rather to investigate whether accurate reconstruction can emerge from purely symbolic structures. In doing so, we explore an alternative representational paradigm that remains largely unexamined in the jigsaw literature.

\textbf{Symbolic Reassembly.}
After tokenization, each puzzle piece is represented by a short sequence of symbols from a fixed vocabulary $V$. The Seq2Seq solver operates exclusively on these tokens, \textit{with no additional direct access to the original pixel values}. To clarify this claim, suppose that all very-bright patches are clustered around centroid 100. During training, this integer is passed to our Seq2Seq solver and mapped by its embedding layer to a learned vector $e_{100}$. This embedding is optimized solely with respect to the reconstruction objective, and is thus not constrained to preserve visual information about the original patch. Consequently, the geometry of the learned representation space is shaped by reconstruction compatibility instead of perceptual proximity (\eg, visually dissimilar patches may become close if they frequently co-occur along compatible borders). This differs fundamentally from continuous approaches such as ViTs, where patches are linearly projected into continuous embeddings, and are hence entangled to visual appearance (visualization available in the Supp.). 

Thus, although pieces are quantized by $q_\phi$ according to pixel values, the Seq2Seq solver $s_\theta$ cannot retrace appearance from its input. This setting places our model in an interesting position: it must infer spatial relationships between pieces using only symbolic representations of their borders. Any notion of compatibility between pieces must therefore emerge from statistical regularities in token co-occurrence and sequence structure, rather than from direct visual cues.

\textbf{Train–Test Domain Alignment.}
A common challenge in continuous feature representations is distributional drift: small photometric variations at test time can produce embeddings that fall outside the distribution seen during training. In contrast, a fixed vocabulary enforces strict alignment between train  and test domains at the symbolic level. At inference time, the model encounters only tokens from the same predefined set $V$. \Ie, although token frequencies may vary, the support of the input space remains unchanged, simplifying generalization.

\textbf{Reduced Variance Through Quantization.}
Our many-to-one quantization scheme maps visually similar patches to the same token, collapsing minor photometric differences that are often irrelevant for border compatibility. From a statistical perspective, this acts as an implicit regularizer by limiting representational resolution and reducing sensitivity to noise. The Seq2Seq model is encouraged to learn structural and combinatorial relationships between pieces rather than relying on fine-grained pixel distinctions.

\textbf{Improved Computational Complexity.}
Scalability is a major concern for modern learning-based jigsaw solvers. Representing each piece by $b$ boundary tokens, the entire puzzle can be encoded using $bN+(N-1)$ integers instead of $N\cdot H\cdot W\cdot C$ pixel values, substantially reducing memory requirements. Moreover, since tokenization is often performed as a preprocessing step, once a puzzle has been discretized, the original images are no longer required for permutation prediction. This decouples visual preprocessing from symbolic reasoning and enables lightweight deployment of the solver $s_\theta$.

In summary, discrete symbolic representations impose a deliberate abstraction over visual input. While this constrains access to fine-grained pixel information, it also reduces variance, enforces vocabulary consistency between training and testing, and significantly lowers computational requirements. These properties make discrete Seq2Seq-based jigsaw solving an appealing alternative framework worthy of systematic investigation.

\subsection{Sequence-to-Sequence Solver}
Unlike prior transformer-based approaches, we model jigsaw reassembly explicitly as a Seq2Seq task, adopting an encoder–decoder architecture. Compared to encoder-only or decoder-only models, this design offers several advantages for structured jigsaw reconstruction (summarized in~\cref{tab:enc-dec}).   

\begin{table}[t]
    \centering
    \caption{Encoder-decoder models show several features favorable for reassembly.}
    \begin{tabular}{lccc}
        \toprule
        \multirow{2}{*}{Architecture}        & Encoder-Only & Decoder-Only & Encoder+Decoder \\
                                             & \cite{kim2025solving,heck2025solving} & \cite{lyu2025jigsaw,wang2025jigsaw} & (this work) \\
        \midrule
        Global bidirectional context of input& \checkmark   & \xmark       & \checkmark        \\
        Autoregressive prediction process    & \xmark       & \checkmark   & \checkmark        \\
        Distinct input/output distributions  & \xmark       & \xmark       & \checkmark       \\
        \bottomrule
    \end{tabular}
    \label{tab:enc-dec}
\end{table}

\textbf{Global bidirectional context of input.} The encoder processes the entire input through self-attention, capturing bidirectional dependencies among all piece tokens. Since successful reconstruction depends critically on understanding the relationships between all puzzle pieces, such global contextual mechanism is essential. Recent LVLM-based attempts~\cite{lyu2025jigsaw,wang2025jigsaw} also employ transformers, but typically rely on a decoder-only language model conditioned on visual embeddings. Despite large parameter counts and constrained evaluation settings (\eg, very small puzzles or limited solution space), these approaches show limited gains, highlighting the importance of explicitly modeling global input structure rather than treating reconstruction as free-form generation.

\textbf{Autoregressive prediction process.} The decoder generates the output sequence autoregressively, predicting one position at a time while conditioning each step on both the encoded input context and all previously assigned placements. This autoregressive formulation decomposes the combinatorial permutation space into conditional decisions, substantially simplifying inference compared to predicting the entire assignment in a single step. Moreover, conditioning on prior outputs encourages incremental construction of globally consistent solutions and naturally discourages invalid configurations (\eg, assigning two pieces to the same position). Notably, we observe that this desirable behavior emerges organically during training, without requiring architectural changes or specialized decoding. In contrast, encoder-only approaches, such as ViT-based solvers~\cite{kim2025solving,heck2025solving}, must predict all assignments simultaneously, often relying on auxiliary mechanisms or explicitly constrain the number of admissible solutions.

\textbf{Distinct input/output distributions.} Using both an encoder and a decoder enables effective modeling of input and output distributions that differ substantially in content, structure, and length. Seq2Seq architectures are the standard choice for inherently asymmetric tasks (\eg., machine translation, text-to-speech) precisely because they decouple input representation from output generation. In our case, the input consists of tokenized puzzle representations over a vocabulary $V$, whereas the output corresponds to index-wise permutations over grid indices $\{1,\dots,N-1\}$. Moreover, the input length $bN+(N-1)$ is typically several times larger than the output length $N$. This asymmetry makes our Seq2Seq encoder–decoder formulation particularly well suited for tokenized puzzle reassembly. In contrast, decoder-only architectures treat the output as a continuation of the input sequence, implicitly modeling both under a shared distribution. Encoder-only models produce contextualized representations that are entirely derived from the input and preserve its length. We argue that these limitations make both approaches less suited to our task, where a clear distinction between input encoding and output generation is essential.

\textbf{Flexible Backbone Choice.} Finally, we note that our framework is agnostic to the specific implementation of the solver module $s_\theta$, and hence compatible with most standard Seq2Seq architectures. In our experiments, we compare several popular transformer-based backbones, as well as more traditional recurrent neural networks (RNNs), to assess their effect on reconstruction (cf.~\cref{fig:backbones}).

\subsection{Training Objective}
Jigsaw puzzle reconstruction is evaluated using two standard metrics: \textit{absolute accuracy}, which measures the fraction of correctly placed pieces; and \textit{perfect accuracy}, which counts the ratio of entirely solved puzzles:
\begin{align}
    &\mathrm{Absolute}(Y,\hat{Y})=\frac{1}{N}\sum_{i=1}^N\mathds{1}[y_i=\hat{y}_i] \;,
    &\mathrm{Perfect}(Y,\hat{Y})=\prod_{i=1}^N\mathds{1}[y_i=\hat{y}_i] \;.
\end{align}
These metrics are non-differentiable and thus cannot be optimized directly. Nonetheless, the notion of perfect accuracy is equivalent to ``exact match'' from  NLP Seq2Seq~\cite{qi2022rasat}. Hence, we follow standard practice in language modeling and minimize the cross-entropy loss between the predicted and true positions. This loss encourages the model to assign high probability to the correct piece at each decoding step, and improves both metric simultaneously, as $\mathrm{Perf}(Y,\hat{Y})\le\mathrm{Abs}(Y,\hat{Y})$ always holds. At inference, we apply the argmax function over the predicted logits in order to choose the next placement.

\section{Sequence Analysis}

How analogous are puzzle tokens and natural language tokens? To better understand the statistical properties of our tokenized puzzle sequences, we conduct a series of classical analyses inspired by corpus linguistics. We examine tokens from the ImageNet 3$\times$3, JPwLEG-3, and JPwLEG-5 datasets (see \cref{sec:results} for detailed descriptions) with granularity $B=4$, reduced dimension $d=2^{10}$, and a vocabulary size of $k=2^{12}$. 

\begin{figure}[b]
    \centering
    \includegraphics[width=\linewidth]{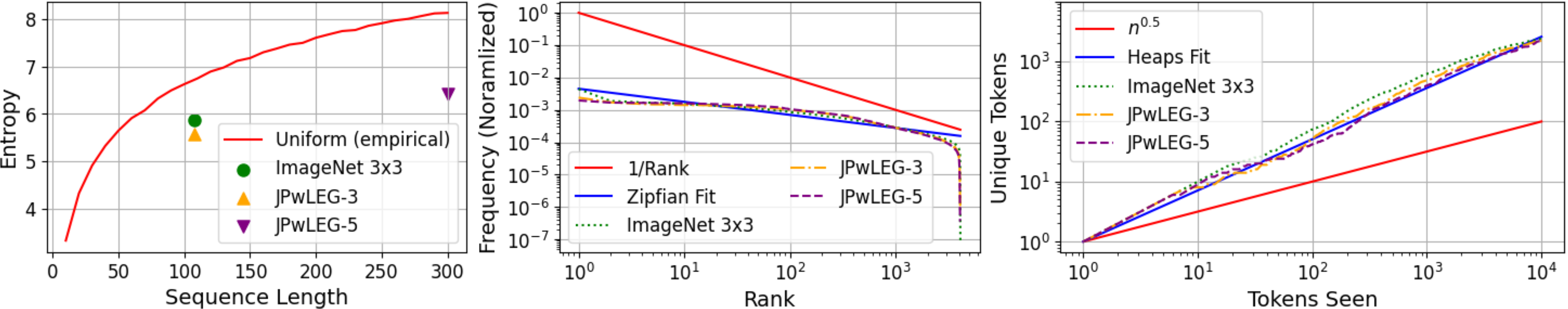}
    \caption{Sequence analysis. We measure the Shannon entropy (left) of our jigsaw tokens alongside their alignment with Zipf's (center) and Heaps' (right) laws.}
    \label{fig:seq_analysis}
\end{figure}

\textbf{Shannon entropy.} Given a discrete random variable $X$ over the token vocabulary $\{0,\dots,|V|-1\}$, its Shannon entropy~\cite{shannon1948mathematical} is defined as:
\begin{equation}
    H(X)=-\sum_{i=0}^{|V|-1}p(X=i)\log p(X=i) \;\;,
\end{equation}
where $p(X=i)$ denotes the probability of observing token $i$. Since the true distribution is unknown, we estimate it using empirical token frequencies. Intuitively, a higher entropy value implies a more random and less predictable sequence. Natural language, for instance, exhibits relatively low entropy due to the uneven distribution of common words and syntactic constraints~\cite{chen2017entropy}.

\cref{fig:seq_analysis} (left) presents the approximated Shannon entropy, averaged over all puzzles, across the three datasets. All entropy scores fall well below the theoretical upper bound of $\log_2(k)=12$, which is expected due to the relatively short sequence lengths ($bN \ll k$). To contextualize these results, we also compare against an empirical baseline derived from uniformly sampling tokens with increasing sequence lengths. Notably, the gap between our datasets and the uniform entropy widens as the number of tokens grows, suggesting that puzzle sequences, like natural language, exhibit a distinguishable structure.

\textbf{Zipf's law.} In natural language, token occurrence often follow a Zipfian distribution: the frequency of a token is inversely proportional to its rank in the frequency table~\cite{powers1998applications}. That is, the most common token occurs roughly twice as often as the second most common, three times as often as the third, and so on. \cref{fig:seq_analysis} (center) plots the empirical frequency–rank curves for our tokenized puzzle sequences, revealing a partially-Zipfian trend. While most tokens follow this distribution, the tail is noticeably sparser than expected under perfect adherence. This might stem from choosing a slightly large $k$ value, which leads to small clusters which are under-represented in the dataset. In~\cref{sec:ablation} we explore the effect of different $k$ values on reconstruction accuracy.

\textbf{Heaps' law.} This property describes how the number of unique tokens grows as a function of sequence length $n$~\cite{heaps1978information}. In natural language, this is typically proportional to $n^\beta$, where $0.4\le\beta\le0.6$. As shown in \cref{fig:seq_analysis} (right), our tokenized puzzles exhibit a steeper curve than the reference line $n^{0.5}$. This suggests that our data display a higher token diversity than typical text, possibly due to the wide variability in patch appearances.

Our analysis reveals both distinct nuances and important structural similarities between puzzle and text tokens, further motivating our language-inspired approach. Interestingly, we have found that the deviations in Zipf's and Heaps' distributions reproduce recent findings regarding image token statistics~\cite{chan2024analyzing}.

\section{Experiments and results}
\begin{figure}[b]
    \centering
    \includegraphics[width=\linewidth]{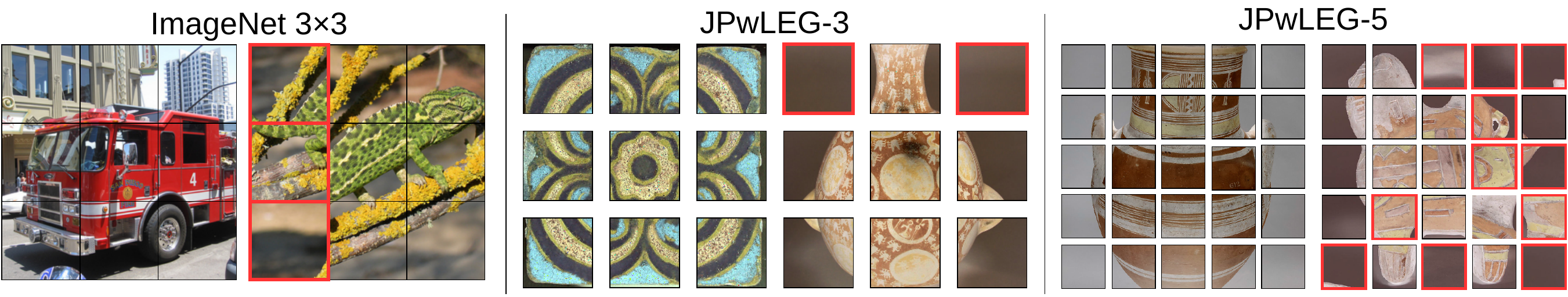}
    \caption{Qualitative reconstruction results. For each datasets we show perfect (left) and partial (right) solutions. Incorrect placements are marked with a red frame.}
    \label{fig:qual}
\end{figure}

\label{sec:results}
We have evaluated our PuzLM over a range of jigsaw reconstruction tasks, both quantitatively and qualitatively, followed by extensive ablation studies. Unless stated otherwise, all experiments were conducted with BART~\cite{lewis2019bart} Seq2Seq backbone, granularity $B=4$, reduced dimension $d=2^{10}$, and vocabulary size $k=2^{12}$. While we focus here on the most widely-used benchmarks, additional experiments (such as cross-dataset generalization, larger puzzles, cost analysis, \etc) as well as full implementation details are provided in the Supp.

\subsection{Puzzle solving benchmarks}
We evaluate our solver across several established puzzle-solving benchmarks. Despite relying solely on tokenized representations, PuzLM achieves strong performance across all datasets, frequently surpassing previous SOTA methods.

\begin{table}[t]
\centering
\begin{minipage}[t]{0.48\textwidth}
\centering
\caption{Results on ImageNet 3$\times$3. Our method achieves new SOTA accuracy.}
\begin{tabularx}{\textwidth}{lcCC}
\toprule
Method    & \# Perm.      & Abs. & Perf. \\
\midrule
CFN~\cite{noroozi2016unsupervised} & \multirow{2}{*}{1000} & 71.0    & NA       \\
Deepzzle~\cite{paumard2020deepzzle}  &                       & 78.6    & 48.5   \\
\midrule
Wei \etal~\cite{wei2019iterative}       & \multirow{4}{*}{9!}   & NA        & 47.3   \\
JPDVT~\cite{liu2024solving}     &                       & 83.3    & 68.7   \\
FCViT~\cite{kim2025solving}     &                       & \underline{90.6}    & \underline{78.9}   \\ 
PuzLM  &                       & \textbf{92.2}                    & \textbf{87.1}        \\
\bottomrule
\end{tabularx}
\label{tab:res-imagenet}
\vspace{6pt}
\caption{Results on ImageNet 3$\times$3 with missing pieces.}
\begin{tabularx}{\textwidth}{lCC|CC|CC}
\toprule
\multirow{2}{*}{Missing} & \multicolumn{2}{c|}{1/9} & \multicolumn{2}{c|}{2/9} & \multicolumn{2}{c}{3/9} \\
              & Abs.       & Perf.      & Abs.       & Perf.      & Abs.       & Perf.      \\
\midrule
JPDVT         & 72.0      & 41.5      & 61.8      & 21.4      & 54.1      & 14.9      \\
PuzLM         & \textbf{86.0} & \textbf{71.3} & \textbf{73.8}  & \textbf{45.1} & \textbf{61.2} & \textbf{23.7} \\
\bottomrule
\end{tabularx}
\label{tab:missing}

\end{minipage}\hfill
\begin{minipage}[t]{0.48\textwidth}
\centering
\caption{
Results on JPwLEG. While remaining competitive on JPwLEG-3, PuzLM outperforms all other solvers on the more challenging JPwLEG-5 benchmark.}
\begin{tabularx}{\textwidth}{lCC|CC}
\toprule
\multirow{2}{*}{Method}         
         & \multicolumn{2}{c|}{JPwLEG-3} & \multicolumn{2}{c}{JPwLEG-5} \\
         & Abs.      & Perf.      & Abs.      & Perf.       \\
\midrule
Deepzzle~\cite{paumard2020deepzzle} & 73.8         & 52.3        & 21.9         & 00.0         \\
SD2RL~\cite{song2023siamese}    & 81.6         & 59.7        & 40.3         & 05.1         \\
PDN-GA~\cite{song2023solving}   & 81.3         & 58.2        & 44.3         & 06.1         \\
ERL-MPP~\cite{song2025erlmpp}  & NA             & NA            & 52.7         & 18.6         \\
DiffAssemble~\cite{scarpellini2024diffassemble}    & 79.7          & 54.2        & 63.0             & 16.4             \\
JPDVT~\cite{liu2024solving}    & 71.3          & NA        & 04.1             & 00.0             \\
VLHSA~\cite{xu2025vlhsa}    & 85.4         & NA        & \underline{66.9}             & \underline{19.0}             \\
FCViT~\cite{kim2025solving}    & \textbf{96.9}         & \textbf{87.9}        & 43.2             & 02.5             \\
PuzLM & \underline{91.9}         & \underline{84.5}        & \textbf{72.1}         & \textbf{32.5}              \\
\bottomrule
\end{tabularx}
\label{tab:res-jpwleg}
\end{minipage}
\end{table}

\textbf{ImageNet 3$\times$3.} 
Studied by some of the first works on data-driven jigsaw puzzle solving~\cite{noroozi2016unsupervised,paumard2020deepzzle}, this benchmark consists of images from the popular ImageNet dataset~\cite{deng2009imagenet}, divided into 3$\times$3 slices. To handle the factorial search space, some methods restrict the task to a fixed-size subset of candidate permutations. We report this number (as \#Perm), and reconstruction accuracy, in \cref{tab:res-imagenet}. Our model outperforms all baselines on ImageNet 3$\times$3, achieving new SOTA reconstruction accuracy. Notably, the relatively large gain in perfect solutions (+$8.2\%$) indicates our strong ability to align global puzzle constraints.

\textbf{JPwLEG.} 
``Jigsaw Puzzles with Large Eroded Gaps'' (JPwLEG) dataset~\cite{song2023siamese} is designed to better resemble real-world fragmentation of archaeological artifacts, by using artwork images from the MET collection~\cite{ypsilantis2021met} with artificially eroded piece boundaries. This fact, combined with a relatively small training set, makes it especially challenging for deep solvers. Results on 3$\times$3 (JPwLEG-3) and 5$\times$5 (JPwLEG-5) puzzles are reported in \cref{tab:res-jpwleg}. On JPwLEG-3, PuzLM achieves a strong 87.9\% perfect accuracy, second only to FCViT~\cite{kim2025solving}, and surpasses multiple models which were specifically designed to handle the JPwLEG erosion~\cite{song2023siamese,song2023solving}. On the larger JPwLEG-5, PuzLM establishes a new SOTA performance, achieving a substantial +13.5\% improvement in perfect solutions compared to the second-best model~\cite{xu2025vlhsa}. This showcases the scalability and robustness of our approach, standing in contrast to various baselines, which excel on the 3$\times$3 puzzles, but struggle to maintain performance in larger 5$\times$5.

\textbf{Missing Pieces.}
A recent work by Liu \etal~\cite{liu2024solving} addressed the challenge of solving jigsaw puzzles with missing pieces. It reflects real-world conditions, where not all pieces are found, making reassembly substantially harder. To adapt our method to this setting, we substitute all tokens corresponding to missing pieces with a dedicated mask token. The prediction protocol remains unchanged, computing accuracy only on the non-masked pieces. Following Liu \etal~\cite{liu2024solving}, \cref{tab:missing} presents results on ImageNet 3$\times$3 with 1, 2, and 3 missing pieces. Across all levels of difficulty, PuzLM achieves higher absolute and perfect accuracy. These gains suggest that, even when pieces are removed, our solver exploits global patterns and semantic consistency to successfully infer plausible solutions.

\subsection{Ablation Studies}
\label{sec:ablation}
Having established PuzLM’s strong performance across diverse benchmarks, we next examine the factors that contribute to its effectiveness. The following ablation studies isolate the influence of our two main modules (image tokenizer and language model) as well as various design choices and hyperparameters, shedding light on how each component supports accurate and scalable reconstruction.

\begin{figure}[t]
    \centering
    \includegraphics[width=\linewidth]{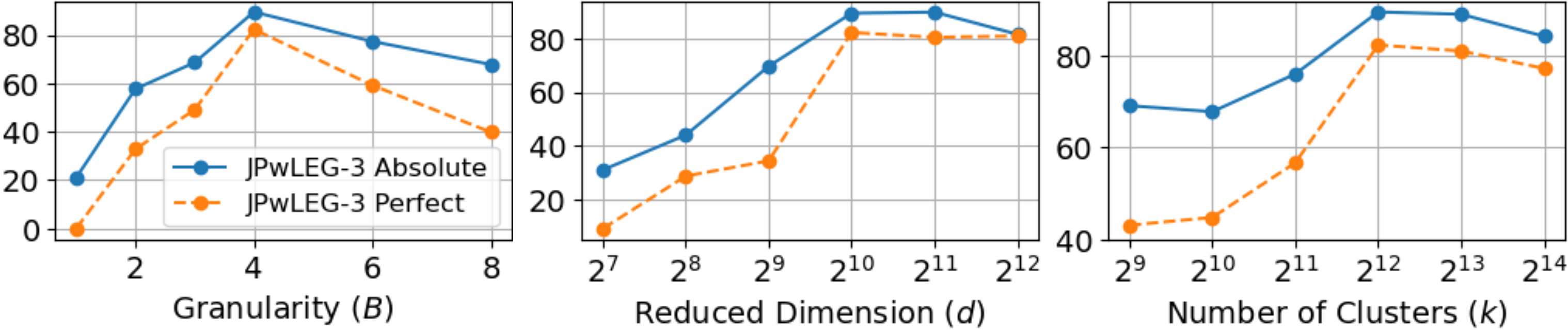}
    \caption{Extensive search of the tokenizer configuration space reveals a common trend: accurate reconstruction stems from a tradeoff between expressivity and abstraction.}
    \label{fig:hyperparams}
\end{figure}

\textbf{Granularity ($B$).}
The granularity of our tokenizer determines the super-token size, and thus plays a central role in the performance of our method. Increasing $B$ leads to a more detailed representation, potentially improving the model’s ability to distinguish between pieces. However, it also increases the sequence length and the complexity of the token relationships, introducing a trade-off between expressiveness and learnability. To evaluate this trade-off, we vary $B$ and report reconstruction accuracy in \cref{fig:hyperparams} (left). Our method shows improved performance as granularity increases, peaking at $B=4$, after which accuracy begins to decline. This suggests that an intermediate granularity level offers the best balance between descriptiveness and length.

\textbf{Reduced Dimension ($d$).} As previously discussed, applying PCA before $k$-means quantization can enhance clustering performance by removing redundant variance from the original data~\cite{mukherjee2024capturing}. However, overly aggressive dimensionality reduction may also eliminate subtle but critical details necessary for distinguishing between different clusters. This trade-off is evident in the experiment shown in \cref{fig:hyperparams} (center), where different $d$ values are evaluated for their effect on JPwLEG-3 accuracy. When the reduced dimension is too small, the representation lacks sufficient detail for effective clustering, while excessively large $d$ values preserve unnecessary variation that also degrades clustering utility.

\textbf{Vocabulary Size ($k$).}
Another important design choice in our tokenizer is the number of centroids used during $k$-means clustering, which determines the vocabulary size. A larger $k$ allows the tokenizer to make finer distinctions between patches, enabling more expressive representations. However, it also increases the number of unique tokens the model must understand, which can introduce sparsity in the dataset and raise architectural requirements. To study this trade-off, we evaluate our model’s performance for increasing $k$. As shown in \cref{fig:hyperparams} (right), reconstruction accuracy improves as $k$ grows, but only up to a point. Beyond a certain vocabulary size, the gains plateau or even diminish, likely due to over-fragmentation of the feature space leading to insufficient token frequency. These results suggest that while a sufficiently expressive vocabulary is essential for reassembly, excessively large token sets might hurt generalization.

\begin{table}[t]
\centering
\begin{minipage}[]{0.69\textwidth}
\centering
\captionof{table}{Alternative tokenizers and design choices.}
\begin{tabularx}{\linewidth}{lCCCCcC}
\toprule
\multirow{2}{*}{Image Tokenizer} & \multirow{2}{*}{Gran.} & \multirow{2}{*}{\shortstack{Vocab.\\Size}} & \multicolumn{2}{c}{JPwLEG-3} & \multirow{2}{*}{\shortstack{Param.\\Count}} & \multirow{2}{*}{\shortstack{Enc.\\Time}} \\
                                 &                                    &                                  & Abs.         & Perf.        & & \\
\midrule
VQ-VAE~\cite{van2017neural}                           & 64                                 & 8192                             &  34.4      &  12.0     & 55M           & 4 ms         \\
\midrule
TiTok-S                          & 128                                & \multirow{3}{*}{4096}            &  16.3        & 04.5         & 84M           & 6 ms          \\
TiTok-B~~\cite{yu2024image}                          & 64                                 &                                  &  36.9       & 10.7      & 205M          & 10 ms         \\
TiTok-L                          & 32                                 &                                  & 41.9         & 20.0        & 641M          & 29 ms         \\
\midrule
PuzLM                         & \multirow{6}{*}{12}         & \multirow{6}{*}{4096}        & \textbf{91.9}          & \textbf{84.5}    & \multirow{6}{*}{$<$1M}          & \multirow{6}{*}{$<$1 ms}              \\
\;\;\; w/o PCA                 &                                    &                                  & 81.1        & 67.0      &             \\
\;\;\; w/o border              &                                    &                                  & 72.2        & 50.3      &             \\
\;\;\; w/o lex. order          &                                    &                                  & 67.6        & 35.0      &             \\
\;\;\; w/o clockwise           &                                    &                                 & \underline{86.9}        & \underline{80.8}         &               \\
\;\;\; w/o sep. token          &                                    &                                  & 84.7        & 79.0      &               \\
\bottomrule
\end{tabularx}
\label{tab:abl-tokenizer}
\end{minipage}
\hfill
\begin{minipage}[]{0.29\textwidth}
\centering
\includegraphics[width=\linewidth]{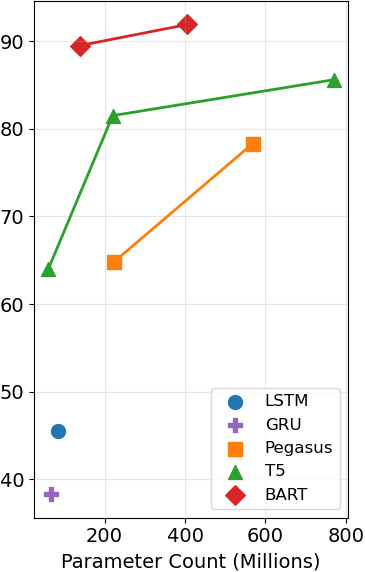}
\captionof{figure}{Experimenting with various Seq2Seq backbones.}
\label{fig:backbones}
\end{minipage}
\end{table}

\textbf{Image Tokenizer.}
Although our proposed tokenizer yields top-tier results, completeness demands proper comparison with deep image quantization approaches, commonly used in vision-language models~\cite{jia2025principles}. We therefore compare against two leading alternatives: the classic Vector-Quantized Variational Autoencoder (VQ-VAE)~\cite{van2017neural}; and the recent TiTok~\cite{yu2024image}, which is known to achieve remarkable image quality using very few tokens. 
\cref{tab:abl-tokenizer} reports the reconstruction accuracy of our model when trained on tokens produced by these alternatives. Despite their strong performance in image generation benchmarks, both VQ-VAE and TiTok underperform in our setting. These methods typically produce longer sequences or entangled representations, which are less suitable for structured tasks like puzzle reassembly. By contrast, our tokenizer is specifically designed to preserve spatial structure in a short, piece-aligned format, resulting in better downstream accuracy and much lower computational cost.

The lower half of \cref{tab:abl-tokenizer} presents an ablation study over key design choices in our tokenizer. Removing the PCA projection (w/o PCA), including non-border tokens (w/o border), or using random rather than lexicographic order (w/o lex. order), each leads to a significant drop in performance, confirming the importance of compact and structured representations. While the decrease in accuracy when replacing the clockwise border traversal with raster scan (w/o clockwise) or dropping the separator token (w/o sep. token) is less dramatic, it is still noticeable. Together, these results highlight the importance of each component in our final tokenizer.

\textbf{Seq2Seq Backbone.}
As previously discussed, our approach is agnostic to the specific implementation of the Seq2Seq architecture. However, as with most machine learning pipelines, this choice plays a critical role in downstream performance. \cref{fig:backbones} compares the JPwLEG-3 absolute reconstruction accuracy of several Seq2Seq models. We evaluate three transformer-based models: BART~\cite{lewis2019bart}, T5~\cite{raffel2020exploring}, and Pegasus~\cite{zhang2020pegasus} (of multiple sizes); as well as traditional LSTM~\cite{hochreiter1997long} and GRU~\cite{cho2014learning} RNNs (see architectural details in the Supp.). Our results show that transformer models significantly outperform the RNN baselines, underscoring the importance of global attention for puzzle reassembly. Among the transformers, BART performs robustly across measures, making it a reliable all-purpose choice. These findings support the idea that architectural scale and context range are especially valuable when solving complex puzzles.

\section{Conclusions}
We present a novel approach to square jigsaw puzzle solving by reframing it as a Seq2Seq prediction task. Our method introduces a lightweight tokenizer that encodes each puzzle piece as a discrete token sequence, enabling standard encoder-decoder language models to reconstruct puzzles. Our approach achieves remarkable results across diverse benchmarks, including degraded, and incomplete puzzles, matching or surpassing vision-centric solvers. These findings highlight the surprising effectiveness of language-inspired representation in a field traditionally dominated by visual methods, suggesting that unconventional perspectives can open new directions for puzzle-solving research.


%
%
\bibliographystyle{splncs04}
\bibliography{main}

\begin{thebibliography}{10}
\providecommand{\url}[1]{\texttt{#1}}
\providecommand{\urlprefix}{URL }
\providecommand{\doi}[1]{https://doi.org/#1}

\bibitem{carion2020end}
Carion, N., Massa, F., Synnaeve, G., Usunier, N., Kirillov, A., Zagoruyko, S.: End-to-end object detection with transformers. In: European conference on computer vision. pp. 213--229. Springer (2020)

\bibitem{chan2024analyzing}
Chan, D.M., Corona, R., Park, J., Cho, C.J., Bai, Y., Darrell, T.: Analyzing the language of visual tokens. arXiv preprint arXiv:2411.05001  (2024)

\bibitem{chang2022maskgit}
Chang, H., Zhang, H., Jiang, L., Liu, C., Freeman, W.T.: Maskgit: Masked generative image transformer. In: Proceedings of the IEEE/CVF conference on computer vision and pattern recognition. pp. 11315--11325 (2022)

\bibitem{chen2017entropy}
Chen, R., Liu, H., Altmann, G.: Entropy in different text types. Digital scholarship in the humanities  \textbf{32}(3),  528--542 (2017)

\bibitem{cho2014learning}
Cho, K., Van~Merri{\"e}nboer, B., Gul{\c{c}}ehre, {\c{C}}., Bahdanau, D., Bougares, F., Schwenk, H., Bengio, Y.: Learning phrase representations using rnn encoder--decoder for statistical machine translation. In: Proceedings of the 2014 conference on empirical methods in natural language processing (EMNLP). pp. 1724--1734 (2014)

\bibitem{deng2009imagenet}
Deng, J., Dong, W., Socher, R., Li, L.J., Li, K., Fei-Fei, L.: Imagenet: A large-scale hierarchical image database. In: 2009 IEEE conference on computer vision and pattern recognition. pp. 248--255. Ieee (2009)

\bibitem{derech2021solving}
Derech, N., Tal, A., Shimshoni, I.: Solving archaeological puzzles. Pattern Recognition  \textbf{119},  108065 (2021)

\bibitem{dosovitskiy2020image}
Dosovitskiy, A., Beyer, L., Kolesnikov, A., Weissenborn, D., Zhai, X., Unterthiner, T., Dehghani, M., Minderer, M., Heigold, G., Gelly, S., et~al.: An image is worth 16x16 words: Transformers for image recognition at scale. arXiv preprint arXiv:2010.11929  (2020)

\bibitem{elkin2025recognizing}
Elkin, G., Shahar, O.I., Ohayon, Y., Alali, N., Ben-Shahar, O.: Recognizing artistic style of archaeological image fragments using deep style extrapolation. In: International Conference on Human-Computer Interaction. pp. 115--131. Springer (2025)

\bibitem{esser2021taming}
Esser, P., Rombach, R., Ommer, B.: Taming transformers for high-resolution image synthesis. In: Proceedings of the IEEE/CVF conference on computer vision and pattern recognition. pp. 12873--12883 (2021)

\bibitem{freeman2006apictorial}
Freeman, H., Garder, L.: Apictorial jigsaw puzzles: The computer solution of a problem in pattern recognition. IEEE Transactions on Electronic Computers (2),  118--127 (1964)

\bibitem{gallagher2012jigsaw}
Gallagher, A.C.: Jigsaw puzzles with pieces of unknown orientation. In: 2012 IEEE Conference on computer vision and pattern recognition. pp. 382--389. IEEE (2012)

\bibitem{gur2017square}
Gur, S., Ben-Shahar, O.: From square pieces to brick walls: The next challenge in solving jigsaw puzzles. In: Proceedings of the IEEE international conference on computer vision. pp. 4029--4037 (2017)

\bibitem{harel2024pictorial}
Harel, P., Shahar, O.I., Ben-Shahar, O.: Pictorial and apictorial polygonal jigsaw puzzles from arbitrary number of crossing cuts. International Journal of Computer Vision  \textbf{132}(9),  3428--3462 (2024)

\bibitem{heaps1978information}
Heaps, H.S.: Information retrieval: Computational and theoretical aspects. Academic Press, Inc. (1978)

\bibitem{heck2025solving}
Heck, G., Lerm{\'e}, N., Le~H{\'e}garat-Mascle, S.: Solving jigsaw puzzles with vision transformers. Pattern Analysis and Applications  \textbf{28}(2), ~110 (2025)

\bibitem{hochreiter1997long}
Hochreiter, S., Schmidhuber, J.: Long short-term memory. Neural computation  \textbf{9}(8),  1735--1780 (1997)

\bibitem{jia2025principles}
Jia, J., Gao, J., Xue, B., Wang, J., Cai, Q., Chen, Q., Zhao, X., Jiang, P., Gai, K.: From principles to applications: A comprehensive survey of discrete tokenizers in generation, comprehension, recommendation, and information retrieval. arXiv preprint arXiv:2502.12448  (2025)

\bibitem{kim2025solving}
Kim, G., Cho, H., Nam, H.: Solving jigsaw puzzles by predicting fragment’s coordinate based on vision transformer. Expert Systems with Applications  \textbf{272},  126776 (2025)

\bibitem{kolesnikov2019revisiting}
Kolesnikov, A., Zhai, X., Beyer, L.: Revisiting self-supervised visual representation learning. In: Proceedings of the IEEE/CVF conference on computer vision and pattern recognition. pp. 1920--1929 (2019)

\bibitem{lewis2019bart}
Lewis, M., Liu, Y., Goyal, N., Ghazvininejad, M., Mohamed, A., Levy, O., Stoyanov, V., Zettlemoyer, L.: Bart: Denoising sequence-to-sequence pre-training for natural language generation, translation, and comprehension. arXiv preprint arXiv:1910.13461  (2019)

\bibitem{li2021jigsawgan}
Li, R., Liu, S., Wang, G., Liu, G., Zeng, B.: Jigsawgan: Auxiliary learning for solving jigsaw puzzles with generative adversarial networks. IEEE Transactions on Image Processing  \textbf{31},  513--524 (2021)

\bibitem{liu2024solving}
Liu, J., Teshome, W., Ghimire, S., Sznaier, M., Camps, O.: Solving masked jigsaw puzzles with diffusion vision transformers. In: Proceedings of the IEEE/CVF Conference on Computer Vision and Pattern Recognition. pp. 23009--23018 (2024)

\bibitem{lyu2025jigsaw}
Lyu, Z., Zhang, D., Ye, W., Li, F., Jiang, Z., Yang, Y.: Jigsaw-puzzles: From seeing to understanding to reasoning in vision-language models. In: Proceedings of the 2025 Conference on Empirical Methods in Natural Language Processing. pp. 26003--26014 (2025)

\bibitem{mukherjee2024capturing}
Mukherjee, C.S., Deorkar, N., Zhang, J.: Capturing the denoising effect of pca via compression ratio. Advances in Neural Information Processing Systems  \textbf{37},  26136--26170 (2024)

\bibitem{noroozi2016unsupervised}
Noroozi, M., Favaro, P.: Unsupervised learning of visual representations by solving jigsaw puzzles. In: European conference on computer vision. pp. 69--84. Springer (2016)

\bibitem{van2017neural}
van~den Oord, A., Vinyals, O., Kavukcuoglu, K.: Neural discrete representation learning. In: Advances in Neural Information Processing Systems. vol.~30, pp. 6306--6315 (2017)

\bibitem{paikin2015solving}
Paikin, G., Tal, A.: Solving multiple square jigsaw puzzles with missing pieces. In: Proceedings of the IEEE conference on computer vision and pattern recognition. pp. 4832--4839 (2015)

\bibitem{paumard2020deepzzle}
Paumard, M.M., Picard, D., Tabia, H.: Deepzzle: Solving visual jigsaw puzzles with deep learning and shortest path optimization. IEEE Transactions on Image Processing  \textbf{29},  3569--3581 (2020)

\bibitem{pomeranz2011fully}
Pomeranz, D., Shemesh, M., Ben-Shahar, O.: A fully automated greedy square jigsaw puzzle solver. In: CVPR 2011. pp. 9--16. IEEE (2011)

\bibitem{powers1998applications}
Powers, D.M.: Applications and explanations of zipf’s law. In: New methods in language processing and computational natural language learning (1998)

\bibitem{qi2022rasat}
Qi, J., Tang, J., He, Z., Wan, X., Cheng, Y., Zhou, C., Wang, X., Zhang, Q., Lin, Z.: Rasat: Integrating relational structures into pretrained seq2seq model for text-to-sql. arXiv preprint arXiv:2205.06983  (2022)

\bibitem{raffel2020exploring}
Raffel, C., Shazeer, N., Roberts, A., Lee, K., Narang, S., Matena, M., Zhou, Y., Li, W., Liu, P.J.: Exploring the limits of transfer learning with a unified text-to-text transformer. Journal of machine learning research  \textbf{21}(140),  1--67 (2020)

\bibitem{rika2025generic}
Rika, D., Sholomon, D., David, E., Pais, A., Netanyahu, N.S.: A generic hybrid framework for 2d visual reconstruction. arXiv preprint arXiv:2501.19325  (2025)

\bibitem{scarpellini2024diffassemble}
Scarpellini, G., Fiorini, S., Giuliari, F., Moreiro, P., Del~Bue, A.: Diffassemble: A unified graph-diffusion model for 2d and 3d reassembly. In: Proceedings of the IEEE/CVF Conference on Computer Vision and Pattern Recognition. pp. 28098--28108 (2024)

\bibitem{shahar2025pairwise}
Shahar, O.I., Elkin, G., Ben-Shahar, O.: Pairwise alignment \& compatibility for arbitrarily irregular image fragments. arXiv preprint arXiv:2507.09767  (2025)

\bibitem{shannon1948mathematical}
Shannon, C.E.: A mathematical theory of communication. The Bell system technical journal  \textbf{27}(3),  379--423 (1948)

\bibitem{sholomon2013genetic}
Sholomon, D., David, O., Netanyahu, N.S.: A genetic algorithm-based solver for very large jigsaw puzzles. In: Proceedings of the IEEE conference on computer vision and pattern recognition. pp. 1767--1774 (2013)

\bibitem{sholomon2016dnn}
Sholomon, D., David, O.E., Netanyahu, N.S.: {DNN}-buddies: A deep neural network-based estimation metric for the jigsaw puzzle problem. In: International Conference on Artificial Neural Networks. pp. 170--178. Springer (2016)

\bibitem{simeoni2025dinov3}
Sim{\'e}oni, O., Vo, H.V., Seitzer, M., Baldassarre, F., Oquab, M., Jose, C., Khalidov, V., Szafraniec, M., Yi, S., Ramamonjisoa, M., et~al.: Dinov3. arXiv preprint arXiv:2508.10104  (2025)

\bibitem{soille2006morphological}
Soille, P.: Morphological image compositing. IEEE Transactions on Pattern Analysis and Machine Intelligence  \textbf{28}(5),  673--683 (2006)

\bibitem{song2023siamese}
Song, X., Jin, J., Yao, C., Wang, S., Ren, J., Bai, R.: Siamese-discriminant deep reinforcement learning for solving jigsaw puzzles with large eroded gaps. In: Proceedings of the AAAI Conference on Artificial Intelligence. vol.~37, pp. 2303--2311 (2023)

\bibitem{song2023solving}
Song, X., Yang, X., Ren, J., Bai, R., Jiang, X.: Solving jigsaw puzzle of large eroded gaps using puzzlet discriminant network. In: ICASSP 2023-2023 IEEE international conference on acoustics, speech and signal processing (ICASSP). pp.~1--5. IEEE (2023)

\bibitem{song2025erlmpp}
Song, X., Yang, X., Yao, C., Ren, J., Bai, R., Chen, X., Jiang, X.: {ERL}-{MPP}: Evolutionary reinforcement learning with multi-head puzzle perception for solving large-scale jigsaw puzzles of eroded gaps. In: Proceedings of the AAAI Conference on Artificial Intelligence. vol.~39, pp. 6968--6977 (2025)

\bibitem{talon2022ganzzle}
Talon, D., Del~Bue, A., James, S.: Ganzzle: Reframing jigsaw puzzle solving as a retrieval task using a generative mental image. In: 2022 IEEE international conference on image processing (ICIP). pp. 4083--4087. IEEE (2022)

\bibitem{talon2025ganzzle++}
Talon, D., Del~Bue, A., James, S.: Ganzzle++: Generative approaches for jigsaw puzzle solving as local to global assignment in latent spatial representations. Pattern Recognition Letters  \textbf{187},  35--41 (2025)

\bibitem{tsesmelis2024re}
Tsesmelis, T., Palmieri, L., Khoroshiltseva, M., Islam, A., Elkin, G., Shahar, O.I., Scarpellini, G., Fiorini, S., Ohayon, Y., Alali, N., Aslan, S., Morerio, P., Vascon, S., gravina, E., Napolitano, M., Scarpati, G., zuchtriegel, G., Sp\"{u}hler, A., Fuchs, M., James, S., Ben-Shahar, O., Pelillo, M., Del~Bue, A.: Re-assembling the past: The repair dataset and benchmark for real world 2d and 3d puzzle solving. Advances in Neural Information Processing Systems  \textbf{37},  30076--30105 (2024)

\bibitem{vardi2023multi}
Vardi, B., Torcinovich, A., Khoroshiltseva, M., Pelillo, M., Ben-Shahar, O.: Multi-phase relaxation labeling for square jigsaw puzzle solving. arXiv preprint arXiv:2303.14793  (2023)

\bibitem{vaswani2017attention}
Vaswani, A., Shazeer, N., Parmar, N., Uszkoreit, J., Jones, L., Gomez, A.N., Kaiser, {\L}., Polosukhin, I.: Attention is all you need. Advances in neural information processing systems  \textbf{30} (2017)

\bibitem{wang2025jigsaw}
Wang, Z., Zhu, J., Tang, B., Li, Z., Xiong, F., Yu, J., Blaschko, M.B.: Jigsaw-r1: A study of rule-based visual reinforcement learning with jigsaw puzzles. arXiv preprint arXiv:2505.23590  (2025)

\bibitem{wei2019iterative}
Wei, C., Xie, L., Ren, X., Xia, Y., Su, C., Liu, J., Tian, Q., Yuille, A.L.: Iterative reorganization with weak spatial constraints: Solving arbitrary jigsaw puzzles for unsupervised representation learning. In: Proceedings of the IEEE/CVF conference on computer vision and pattern recognition. pp. 1910--1919 (2019)

\bibitem{xu2025vlhsa}
Xu, Z., Liu, X.: Vlhsa: Vision-language hierarchical semantic alignment for jigsaw puzzle solving with eroded gaps. arXiv preprint arXiv:2509.25202  (2025)

\bibitem{ypsilantis2021met}
Ypsilantis, N.A., Garcia, N., Han, G., Ibrahimi, S., Van~Noord, N., Tolias, G.: The met dataset: Instance-level recognition for artworks. In: Thirty-fifth conference on neural information processing systems datasets and benchmarks track (Round 2) (2021)

\bibitem{yu2024image}
Yu, Q., Weber, M., Deng, X., Shen, X., Cremers, D., Chen, L.C.: An image is worth 32 tokens for reconstruction and generation. Advances in Neural Information Processing Systems  \textbf{37},  128940--128966 (2024)

\bibitem{zhang2020pegasus}
Zhang, J., Zhao, Y., Saleh, M., Liu, P.: Pegasus: Pre-training with extracted gap-sentences for abstractive summarization. In: International conference on machine learning. pp. 11328--11339. PMLR (2020)

\end{thebibliography}
\end{document}